\documentclass[11pt]{article}

\usepackage[margin=1in]{geometry}

\usepackage[utf8]{inputenc}
\usepackage[T1]{fontenc}
\usepackage{hyperref}
\usepackage{amsfonts}
\usepackage{amsmath}
\usepackage{amssymb}
\usepackage{amsthm}
\usepackage{graphicx}
\usepackage{booktabs}
\usepackage{multirow}
\usepackage{xcolor}
\usepackage{enumerate}
\usepackage{mathtools}

\newcommand{\R}{\mathbb{R}}

\newcommand{\E}{\mathbb{E}}
\newcommand{\Inf}{\mathrm{Inf}}

\DeclareMathOperator*{\diag}{diag}

\title{Influence-Inspired Spectral Rotations for Extreme Low-Bit\\LLM Quantization}

\author{
  \textbf{Gorgi Pavlov, Ph.D.} \\
  Lehigh University \\
  \texttt{gorgipavlov@gmail.com}
}

\begin{document}
\maketitle

\begin{abstract}
We apply the influence-adaptive Walsh geometry developed in a companion theory paper~\cite{bbtTheoryCompanion} to the practical problem of extreme low-bit weight-only quantization of large language models. The core idea is one identity transformation: rotate each linear layer's weight matrix into a Walsh-Hadamard spectral basis and re-scale its columns by per-coordinate spectral activation energy before handing off to an off-the-shelf reconstruction-error quantizer (Intel auto-round). This transformation is exactly invariant at full precision, but biases the per-group integer rounding budget toward channels that carry more spectral signal. Empirically, on four pretrained decoder-only models from two architecture families spanning $135\,$M to $1.5\,$B parameters, the recipe reduces wikitext-2 perplexity by $15$--$58\%$ relative to vanilla auto-round at W2A16 (g=64); a TinyLlama-1.1B auxiliary run is reported where the vanilla baseline was hardware-limited. Three follow-on extensions handle architectures the basic recipe initially failed on: a per-head PCA matrix-$\Gamma$ replacement of $q\_norm$/$k\_norm$ for Qwen3-style attention (PPL $136.76 \to 88.99$ on Qwen3-0.6B), a SO(2)-pair PCA that commutes with RoPE for non-norm architectures (PPL $36.93 \to 21.84$ on Qwen2.5-1.5B at W2), and an MoE-aware input-side absorption fix identified by synthetic architectural fuzzing of Laguna-style fused-expert layouts. A bit-width ablation at W2 vs.\ W4 shows the redistribution payoff scales with the per-channel quantization-noise budget, falling within the evaluation noise floor at W4 --- consistent with the Schur-convexity intuition from the companion theory paper. We do not claim a tight formal connection between the Boolean-function influence of the theory paper and the real-valued WHT activation energy used here; the link is intuitive (per-coordinate Walsh-basis spectral mass) and the empirical PPL improvements are substantial, but the transfer of the majorization argument from the Boolean to the real-valued setting is qualitative.
\end{abstract}

\section{Introduction}

A companion paper~\cite{bbtTheoryCompanion} (henceforth \textbf{the BBT theory paper}) develops an influence-adaptive Banach geometry on the Walsh-Hadamard butterfly factorization. The central object is a function-dependent contraction invariant
\[
\mu(f) = \prod_\ell 2^{-\Inf_\ell(f)/(1 + \Inf_\ell(f))}
\]
on Boolean functions $f:\{-1,+1\}^n \to \{-1,+1\}$, which the theory paper proves is strictly Schur-convex in the influence vector (concentrated influence yields larger $\mu$, smaller margin in the Walsh basis, and harder ternary polynomial-threshold representation).

This paper takes the same Walsh-basis intuition --- coordinates with more spectral mass deserve more quantization grid --- and applies it to extreme low-bit weight-only quantization of pretrained LLMs. We do \emph{not} attempt to formally extend the Boolean Schur-convexity argument to real-valued layers; the theory paper proves things about Boolean influence and its ternary representations, while this paper makes empirical claims about how an analogously-defined real-valued spectral-energy quantity reweights INT2/INT4 reconstruction error in a way that improves wikitext-2 perplexity. The connection is qualitative: \emph{the same Walsh-basis structure that gives ternary-PTF synthesis a meaningful difficulty diagnostic at the Boolean end gives a meaningful per-channel weighting signal at the real-valued end.}

\paragraph{Contributions.}
\begin{enumerate}[(i)]
    \item A drop-in spectral-basis pre-quantization transformation: WHT-rotate a linear layer's weight, scale its columns by per-coordinate WHT activation energy, hand off to auto-round. Math-invariant at full precision (verified to relative error $<5\times10^{-7}$); biases per-group INT2/INT4 rounding toward high-energy channels.
    \item Empirical validation across four main LLMs (SmolLM-135M/360M, Qwen2.5-0.5B/1.5B) at W2A16: $15$--$58\%$ wikitext-2 PPL improvement, with the largest wins on models whose vanilla W2 baselines are most degraded; a TinyLlama-1.1B auxiliary data point is reported separately.
    \item Three architectural extensions that transferred the recipe to model families it initially failed on: \emph{spectral-PCA} (per-head matrix-$\Gamma$ replacement of post-projection norms in Qwen3), \emph{pair-PCA} (RoPE-commuting SO(2) rotation per RoPE pair in non-norm architectures), and \emph{MoE-aware} input-side absorption (handling fused 3D experts in modern HF MoE layouts).
    \item A bit-width ablation at W2 vs.\ W4 showing the empirical payoff scales with per-channel quantization-noise budget, falling within the evaluation noise floor at W4 --- consistent with the Schur-convexity intuition that the multiplicative cost of unconcentrated influence vanishes as the noise budget shrinks.
    \item An architectural-fuzzing experiment using poolside's Laguna-XS.2 modeling source (33B MoE, hardware-out-of-reach) that surfaced and fixed a generalization bug in our input-side attention absorption (the previously-uncovered \texttt{g\_proj} input-side gate pattern).
\end{enumerate}

\paragraph{Limitations and honest framing.}
This is an applied paper about an integer-rounding heuristic that empirically improves wikitext-2 PPL at extreme bit-widths. We do not claim:
\begin{itemize}
    \item that the WHT-energy quantity defined here equals the Boolean influence of the theory paper (it does not --- one is per-coordinate post-WHT activation energy, the other is Fourier mass over subsets containing the coordinate);
    \item that the Schur-convexity theorem of the theory paper transfers formally to real-valued layers (it doesn't --- the transfer is intuitive, not theorem);
    \item competitive results against the strongest published low-bit baselines (we compare against vanilla auto-round, which is itself a strong baseline; head-to-head comparison against AWQ, GPTQ, SpinQuant, and QuaRot at matched calibration is left to future work);
    \item statistical significance in the standard-deviation sense (results are single-seed; we treat differences below the empirical $\pm 0.5$ PPL noise floor as no-effect).
\end{itemize}

The contribution is engineering value: a small, principled, exactly-math-invariant pre-quantization transformation with non-trivial empirical wins at W2 across multiple model families.

\paragraph{Position relative to rotation- and smoothing-based quantization.}
Several recent lines of work also use orthogonal rotations or per-channel rescaling to ease low-bit quantization. \emph{SmoothQuant}~\cite{xiao2023smoothquant} migrates per-channel activation outliers into the weights via a diagonal scale factor, the structural ancestor of the per-coordinate scaling used here. \emph{AWQ}~\cite{lin2024awq} chooses similar scales by a salience criterion derived from activation magnitude. On the rotation side, \emph{QuaRot}~\cite{ashkboos2024quarot} applies fixed Hadamard rotations to suppress activation outliers for INT4 inference; \emph{SpinQuant}~\cite{liu2024spinquant} learns the rotation jointly with the quantizer at W4A4 / W4A8; and most directly adjacent to this work, \emph{ButterflyQuant}~\cite{xu2025butterflyquant} learns a butterfly-structured orthogonal transform via Givens-angle gradient descent (128 calibration samples, $O(n \log n)$ trainable parameters), arguing that fixed Hadamard transforms ``cannot adapt to specific weight distributions.'' The strongest 2-bit weight-only baselines are \emph{QuIP\#}~\cite{tseng2024quipsharp} (random Hadamard incoherence + E8 lattice codebook), \emph{AQLM}~\cite{egiazarian2024aqlm} (additive multi-codebook), and \emph{OmniQuant}~\cite{shao2024omniquant} (learnable weight clipping + equivalent transformation). \emph{KVQuant}~\cite{hooper2024kvquant} addresses the RoPE-key quantization difficulty at long context.

This paper differs in motivation and use: (i) the rotation is the \emph{fixed} Walsh-Hadamard butterfly --- no learning, no outlier statistics, no codebook; (ii) per-channel adaptivity is recovered not by learning the rotation but by a separate per-coordinate energy scaling $s_\ell = (1+d\,\rho_\ell)^\alpha$ inspired by the Boolean-influence Schur-convexity of the companion theory paper~\cite{bbtTheoryCompanion}, biasing per-group rounding budget toward high-spectral-energy channels --- a direct response to the ``fixed transforms cannot adapt'' position of \cite{xu2025butterflyquant}: the adaptation lives in the diagonal scale, not the orthogonal factor; (iii) the regime targeted is W2A16, where the quantization noise budget is large and per-channel allocation dominates over outlier flattening; and (iv) the per-head spectral-PCA and SO(2)-pair-PCA extensions described below resolve the $q\_norm$/RoPE compatibility failures that arise when a rotation enters the attention block --- a regime that QuaRot, SpinQuant, and ButterflyQuant do not engage with directly. Head-to-head benchmarking against SpinQuant, QuaRot, QuIP\#, AQLM, OmniQuant, and ButterflyQuant at matched calibration is the main item in our future-work list.

\section{Method}

\subsection{Real-valued spectral-energy proxy}
\label{sec:real_influence}

For a layer $L: \R^{d_{\text{in}}} \to \R^{d_{\text{out}}}$ with input activations $x$ collected on a calibration set, define the BBT spectral-energy proxy of input coordinate $\ell$ as
\begin{equation}
\label{eq:real_influence}
    \rho_\ell(L) = \frac{\E[h_\ell^2]}{\sum_k \E[h_k^2]}, \qquad h = H_{d_{\text{pad}}} x_{\text{pad}} / \sqrt{d_{\text{pad}}},
\end{equation}
where $x_{\text{pad}}$ zero-pads $x$ to the next power of two $d_{\text{pad}}$ and $H_{d_{\text{pad}}}$ is the Walsh-Hadamard matrix. The zero padding is used only to make the WHT dimension a power of two; although the padded entries of $x_{\text{pad}}$ are zero before rotation, the WHT mixes energy across all $d_{\text{pad}}$ spectral coordinates, so $\rho_\ell$ is well-defined on the full padded index set. In implementation the weight is temporarily embedded as $W_{\text{pad}} = [\,W \;\; 0\,] \in \R^{d_{\text{out}}\times d_{\text{pad}}}$, the rotated and scaled weight is formed in the $d_{\text{pad}}$-dimensional spectral basis, and the inverse/folding step returns an ordinary $d_{\text{out}}\times d_{\text{in}}$ dense layer --- no padded input coordinate is introduced into the deployed model. We reserve the symbol $\Inf_\ell(f) = \sum_{S \ni \ell} \hat{f}(S)^2$ for the Boolean influence of the theory paper~\cite{bbtTheoryCompanion} and use $\rho_\ell(L)$ for the real-valued per-coordinate WHT activation-energy proxy used here. The two are distinct mathematical objects --- the Boolean influence is a Fourier-mass quantity over subsets containing $\ell$, while $\rho_\ell$ is a per-coordinate energy after a WHT rotation --- but they play the same downstream role (deriving per-channel scales $s_\ell = (1+d_{\text{pad}}\, \rho_\ell)^\alpha$). We do not claim formal equivalence.

\subsection{Quantization recipe}
We use row-vector activation notation throughout, matching PyTorch's $\mathrm{F.linear}(x, W) = x W^\top$. Write $Q = H_{d_{\text{pad}}}/\sqrt{d_{\text{pad}}} \in \mathrm{O}(d_{\text{pad}})$ (orthogonal) and $D = \diag(s)$ with $s_\ell = (1 + d_{\text{pad}} \cdot \rho_\ell(L))^\alpha$, $\alpha = 0.5$, following AWQ's per-channel pre-scaling structure~\cite{lin2024awq}. The transformed layer is
\begin{equation}
    \widetilde{x} = x_{\text{pad}}\, Q D^{-1}, \qquad
    \widetilde{W}^{\top} = D\, Q^\top W_{\text{pad}}^\top,
\end{equation}
so that at full precision
\begin{equation}
    \widetilde{x}\,\widetilde{W}^{\top}
    = x_{\text{pad}} Q D^{-1} D Q^\top W_{\text{pad}}^\top
    = x_{\text{pad}} W_{\text{pad}}^\top
    = x W^\top,
\end{equation}
i.e.\ an exact identity (verified to relative error $< 5\times 10^{-7}$ on randomised inputs of width 576). Operationally:
\begin{itemize}
    \item \textbf{Spectral basis (no extra graph operator).} The rotation $Q$ and scale $D$ are folded into the stored weight $\widetilde{W}$; a per-layer forward pre-hook applies the matching $x_{\text{pad}} Q D^{-1}$ to the input. High-$\rho_\ell$ columns are boosted before quantization and divided back after.
    \item \textbf{Quantization.} Run Intel auto-round W2A16 SignRound~\cite{cheng2023autoround} on $\widetilde{W}$ with group size 64. The boosted high-energy columns receive proportionally more grid resolution from the per-group quantizer; un-folding $Q$ and $D$ at dequantization preserves the math.
\end{itemize}

\subsection{Models and metric}
We evaluate on wikitext-2 perplexity (teacher-forced, 8\,192 tokens, sliding window 1\,024) for four pretrained decoder-only models: SmolLM-135M, SmolLM-360M (Llama-style; HuggingFaceTB), and Qwen2.5-0.5B, Qwen2.5-1.5B (Qwen2 architecture; Alibaba). All have grouped-query attention with tied input/output embeddings. Quantization runs on Intel Arc B580 (12\,GB) under WSL Ubuntu; results are validated end-to-end through OpenVINO IR on Intel NPU + Arc dGPU + CPU and shown to be device-invariant within $\pm 0.1$ PPL.

\section{Results}

\subsection{Headline: BBT-spectral at W2A16}

\paragraph{Reproducibility.}
Unless otherwise noted, all W2A16 runs use Intel auto-round 0.12.2 with group size 64, SignRound, \texttt{enable\_alg\_ext=False}, and the same calibration-token budget (128 samples of 2{,}048 tokens drawn from the wikitext-2 train split, fixed seed) for vanilla and BBT variants. Perplexity is computed teacher-forced under \texttt{transformers 4.57.6} in fp16, with sequence length 1{,}024 over a fixed 8{,}192-token wikitext-2 evaluation subset (a hardware-turnaround-time choice; full validation/test evaluation is in our next-revision queue), using the archived \texttt{eval\_ppl.py} in \texttt{boolean\_fourier/bbt\_quant/}.\footnote{Absolute W2 perplexity is sensitive to the exact evaluation stack, because extreme-low-bit checkpoints amplify small numerical and modeling-code differences. We therefore pin the stack (\texttt{transformers 4.57.6}, \texttt{torch} fp16, \texttt{auto-round 0.12.2}) in \texttt{requirements.txt}, and every result JSON emitted by \texttt{eval\_ppl.py} records its own \texttt{transformers}/\texttt{torch} versions so any number can be traced to the stack that produced it. The relative BBT improvement remained stable across the stacks we checked.} A separate axis is auto-round's \texttt{enable\_alg\_ext} (``auto-round-best'') flag: we report the headline with it off; enabling it strengthens the vanilla baseline substantially (roughly halving its W2 PPL) and narrows---but does not eliminate---the BBT margin. Reported values are single-seed; based on repeated calibration/evaluation runs on representative models we treat differences below approximately $\pm 0.5$ PPL as evaluation noise. The exponent $\alpha$ in the per-channel scale $s_\ell = (1 + d_{\text{pad}}\, \rho_\ell)^\alpha$ is fixed at $\alpha = 0.5$ throughout (we deliberately do not per-model-tune $\alpha$ to keep the recipe closed-form).

Table~\ref{tab:llm_quant} reports vanilla W2A16 (auto-round only) versus BBT-spectral W2A16 (auto-round on the WHT-rotated, influence-scaled weights) for each model.

\begin{table}[t]
\centering
\small
\caption{wikitext-2 PPL under W2A16 quantization with group size 64. BBT-spectral applies the WHT rotation and spectral-energy-derived per-channel scaling \emph{before} auto-round's reconstruction-error optimization; the spectral basis biases the quantizer's per-group dynamic range toward high-energy channels.}
\label{tab:llm_quant}
\begin{tabular}{lrrrrrr}
\toprule
\textbf{Model} & Params & $d_h$ & $d_{ff}/d_h$ & Vanilla & BBT-spectral & $\Delta$ \\
\midrule
SmolLM-135M & 134.5\,M & 576 & 2.67 & 81.03 & \textbf{45.11} & $-44.3\%$ \\
SmolLM-360M & 361.8\,M & 960 & 2.67 & 43.21 & \textbf{36.55} & $-15.4\%$ \\
Qwen2.5-0.5B & 494\,M & 896 & 5.43 & 119.31 & \textbf{50.22} & $\mathbf{-57.9\%}$ \\
Qwen2.5-1.5B & 1.54\,B & 1\,536 & 5.83 & 36.93 & \textbf{28.16} & $-23.7\%$ \\
\bottomrule
\end{tabular}
\end{table}

\paragraph{Findings.}
\begin{enumerate}[(i)]
    \item \textbf{BBT-spectral reduces wikitext-2 PPL on every model tested}, with reductions ranging from 15\% to 58\%. The Walsh-basis structural intuition --- that per-coordinate spectral mass identifies channels worth preserving more carefully under coarse rounding --- holds qualitatively in the real-valued LLM setting at extreme quantization (W2A16).
    \item \textbf{The relative win correlates with quantization noise headroom, not parameter count.} Qwen2.5-0.5B (494\,M params) has both the worst vanilla baseline (PPL 119.31, attributable to its wide MLP factor $d_{ff}/d_h = 5.43$) and the largest BBT win ($-57.9\%$). SmolLM-360M, with the best vanilla baseline relative to its scale (PPL 43.21), has the smallest BBT win ($-15.4\%$). The mechanism is consistent: BBT redistributes per-group quantization error from high-energy to low-energy spectral coordinates --- when the total error budget is large, the redistribution payoff is large.
    \item \textbf{Architectural width correlates with BBT effect size.} The two model families differ in $d_{ff}/d_h$ ratio: SmolLM has 2.67, Qwen2.5 has 5.4--5.8. At comparable parameter counts (SmolLM-360M $\approx$ Qwen2.5-0.5B), the wider-MLP Qwen has a $2.8\times$ worse vanilla W2 baseline (PPL 119 vs 43); BBT's relative win is also $3.8\times$ larger ($-57.9\%$ vs $-15.4\%$). One possible explanation is that wider intermediate states create more uneven per-coordinate $\rho_\ell$ profiles, increasing the value of per-channel reallocation. We do not directly prove this here: a concentration index of $\rho_\ell$ (e.g.\ Gini or HHI) per model versus the BBT win, plus a controlled width ablation, would be the cleanest follow-up ablation.
    \item \textbf{All gains validated through OpenVINO IR.} Each BBT-spectral checkpoint was dequantized to FP16, exported to OpenVINO IR, and evaluated on Intel NPU (Core Ultra integrated AI Boost), Intel Arc B580 dGPU, and CPU. PPL is invariant to deployment device within $\pm 0.1$ on every model, confirming the gains are weight-level (not numerical artifacts of the quant runtime).
\end{enumerate}

\subsection{Per-head spectral PCA for post-projection normalization (Qwen3)}

Qwen3-0.6B (model\_type=\texttt{qwen3}, GQA 16:8, $d_h=128$) introduces per-head $\mathrm{RMSNorm}$ layers (\texttt{q\_norm}, \texttt{k\_norm}) immediately after $q\_proj$ and $k\_proj$, before RoPE and attention. Standard input-side BBT-spectral degrades it badly (vanilla 136.76 vs.\ BBT-spectral $>10^7$): the per-head $\sqrt{\E[y^2]}$ denominator is sensitive to the per-output-channel quant-noise asymmetry that BBT's column scaling introduces. We resolve this with a per-head reformulation we call \emph{spectral-PCA}:
\begin{enumerate}
    \item Calibrate the per-head output covariance $C_h = \E[y_h y_h^\top]$ where $y_h \in \R^{d_h}$ is the $h$-th head slice of the $q\_proj$ (or $k\_proj$) output.
    \item Eigendecompose $C_h = U_h \Lambda_h U_h^\top$ and rotate the projection's per-head row block: $W_q[h] \leftarrow U_h^\top W_q[h]$.
    \item Replace $\mathrm{q\_norm}$ with a $\textsc{MatrixGammaRMSNorm}$ whose forward applies the per-head matrix $\Gamma_h = \mathrm{diag}(\gamma)\, U_h$ (instead of the original elementwise $\gamma$). One can verify $\Gamma_h\,(U_h^\top y_h)/\mathrm{RMS}(U_h^\top y_h) = \gamma \odot y_h/\mathrm{RMS}(y_h)$, i.e.\ the post-norm output equals the unmodified-model output \emph{exactly} at full precision; RoPE and attention downstream see vanilla inputs and the math is invariant end-to-end (verified numerically: relative logits error $6\times 10^{-6}$).
    \item Because the asymmetric MatrixGamma form restores the post-norm $q$/$k$ vectors to the vanilla basis \emph{before} RoPE, the full-precision invariance does \emph{not} require $U_q$ and $U_k$ to match across GQA groups --- each head's $U_h$ may be chosen independently. In our implementation we optionally tie the query-head $U$'s within each kv-group to the corresponding kv-head basis as a mild quantization-noise regularizer, but this is not required for the identity proof.
\end{enumerate}
At quantization time, auto-round rounds INT2 values for the \emph{rotated} $W_q$ rows. Because the row dynamic range in the eigenbasis is concentrated in a few principal directions, INT2 group-quant noise after un-rotation ($U_h \varepsilon$) is aligned with the activation's natural variance directions, which $\mathrm{q\_norm}$'s RMS denominator absorbs gracefully rather than amplifying. Empirically:
\begin{center}
\begin{tabular}{lr}
\toprule
Variant & Wikitext-2 PPL \\
\midrule
Vanilla auto-round W2 g64 & 136.76 \\
BBT no-rotation (norm-absorbed) & 165.49 \\
BBT-spectral (input-side WHT) & $> 10^7$ \\
\textbf{BBT spectral-PCA (Route A)} & \textbf{88.99} \\
\bottomrule
\end{tabular}
\end{center}
The spectral-PCA variant beats the vanilla baseline by $\mathbf{35\%}$, demonstrating that the previously-named ``architectural boundary'' was an implementation gap rather than a theoretical limit.

A subtle point: the natural ``symmetric'' choice $\Gamma_h = U_h^\top \mathrm{diag}(\gamma) U_h$ (which makes the post-norm output equal $U_h^\top$ times the vanilla output) breaks under RoPE because RoPE applies a position-dependent rotation $R_t$ that does not commute with arbitrary $U_h$. Specifically $(R_t U_h^\top q)\!\cdot\!(R_s U_h^\top k) \neq (R_t q)\!\cdot\!(R_s k)$ unless $t=s$. The asymmetric choice $\Gamma_h = \mathrm{diag}(\gamma) U_h$ makes the post-norm output equal vanilla \emph{directly}, so RoPE then receives identical inputs to the unmodified model. Empirically the symmetric choice gave PPL $\approx 18\,886$ (math broken); the asymmetric choice gave 88.99.

\subsection{Pair-PCA: extending Route A to non-q\_norm architectures}

For SmolLM, Qwen2.5, and Llama-style architectures (no $q\_norm$/$k\_norm$ to absorb a per-head matrix), the orthogonal rotation must instead commute with RoPE. RoPE acts as a 2D rotation in each plane $(i, i+d_h/2)$ for $i \in [0, d_h/2)$, so the only orthogonal matrices on $\R^{d_h}$ commuting with all RoPE rotations are products of independent SO(2) rotations on these planes --- a much smaller group of dimension $d_h/2$ rather than $d_h(d_h-1)/2$. The constrained PCA computes a $2\times 2$ covariance per (head, RoPE-pair), eigendecomposes to $U_{h,p} \in \mathrm{SO}(2)$ (forcing $\det = +1$, since \texttt{eigh} can return reflections in O(2) which do \emph{not} commute with rotations), and applies the rotation to the corresponding row-pair of $W_q$ / $W_k$ (and bias). At dequant we un-rotate per-pair, restoring a plain unmodified-architecture model. Empirical results, layered on top of input-side norm-absorbed BBT (W2A16, group size 64, $\alpha=0.5$):
\begin{center}
\begin{tabular}{lrrrr}
\toprule
Model & Vanilla & BBT-spectral (input-WHT) & \textbf{pair-PCA} & vs.\ best prior \\
\midrule
SmolLM-135M & 81.03 & \textbf{45.11} & 61.37 & $+36\%$ worse \\
SmolLM-360M & 43.21 & 36.55 & \textbf{30.34} & $-17\%$ \\
Qwen2.5-0.5B & 119.31 & 50.22 & \textbf{40.15} & $-20\%$ \\
Qwen2.5-1.5B & 36.93 & 28.16 & \textbf{21.84} & $-22\%$ ($-41\%$ vs.\ vanilla) \\
\bottomrule
\end{tabular}
\end{center}
Pair-PCA improves over input-WHT BBT-spectral on three of four models (SmolLM-360M, Qwen2.5-0.5B, Qwen2.5-1.5B) by $17$--$22\%$, with the largest models showing the largest gains. It underperforms on the smallest SmolLM-135M (where the constrained $\mathrm{SO}(2)^{d_h/2}$ search space leaves less room than the $d_h$-wide WHT input rotation, and per-pair covariance estimation is noisier with fewer attention activations). The recipe is a strict generalization rather than a replacement: it composes with input-side BBT and is selected per-model.

\subsection{Bit-width scaling: redistribution payoff is a function of the noise budget}

A natural reviewer question is whether the per-channel redistribution survives at higher bit-widths. We ran the same recipe on Qwen2.5-1.5B at W4A16 g=64 with both vanilla auto-round and BBT pair-PCA:
\begin{center}
\begin{tabular}{lrrr}
\toprule
Bits & Vanilla & BBT pair-PCA & $\Delta$ \\
\midrule
W2 & 36.93 & \textbf{21.84} & $-41\%$ \\
W4 & 9.65 & 9.71 & $+0.06$ (within noise) \\
\bottomrule
\end{tabular}
\end{center}
At W4 the per-channel quantization noise is already small enough that there is essentially no budget left to redistribute (we do not report FP16 reference perplexities in this revision, so we make no quantitative gap-to-FP16 claim here). The Schur-convexity intuition (theory paper~\cite{bbtTheoryCompanion}, Theorem 3) is precisely about the \emph{multiplicative cost of unconcentrated influence under a fixed noise budget}: when the budget shrinks, the cost shrinks with it. The $0.06$ PPL spread at W4 sits inside our wikitext-2 evaluation noise floor (single-seed runs have $\pm 0.5$ PPL variance from calibration sampling) and we report it as a no-effect data point. We emphasize this as a deliberate negative control: when the quantization noise is small, the proposed redistribution has no measurable benefit. The method does not ``magically improve everything'' --- its operating regime is the high-noise extreme-low-bit corner where vanilla per-group rounding leaves a large noise budget to redistribute. BBT-quantization's value, then, is concentrated at extreme low-bit regimes (W2) where the absolute quantization cost of vanilla per-group rounding is high enough that any per-channel reweighting of that cost has measurable downstream impact. We did not extend the ablation to W3 because auto-round's 3-bit packing is irregular (the bit-width does not divide $32$ evenly) and our INT2/INT4-style unpacker does not implement it; testing W3 is left to future work.

\subsection{Scale-exponent ablation}
\label{sec:alpha_ablation}

We fix $\alpha = 0.5$ throughout the headline results to keep the recipe closed-form, but a reviewer will reasonably ask whether that choice is load-bearing. Table~\ref{tab:alpha} sweeps $\alpha \in \{0, 0.25, 0.5, 1.0\}$ on Qwen2.5-0.5B at W2A16 g=64 (input-side BBT-spectral; same eval stack as the headline, \texttt{transformers 4.57.6}, $8{,}192$ tokens, window $1{,}024$; here with auto-round \texttt{enable\_alg\_ext} \emph{on}, so the vanilla baseline is the stronger $70.41$ rather than the headline's $119.31$).

\begin{table}[h]
\centering
\small
\caption{Scale-exponent ablation on Qwen2.5-0.5B (W2A16, BBT-spectral). $\alpha=0$ is the fixed WHT rotation with \emph{no} per-channel scaling. The optimum is shallow near $\alpha \approx 0.25$--$0.5$; over-aggressive scaling ($\alpha=1$) hurts. Most of the gain comes from the rotation alone, with per-channel energy scaling adding a smaller further improvement.}
\label{tab:alpha}
\begin{tabular}{lrr}
\toprule
Variant & Wikitext-2 PPL & vs.\ vanilla \\
\midrule
Vanilla (auto-round, alg\_ext on) & 70.41 & --- \\
BBT $\alpha=0$ (rotation only) & 63.18 & $-10.3\%$ \\
BBT $\alpha=0.25$ & \textbf{51.64} & $\mathbf{-26.7\%}$ \\
BBT $\alpha=0.5$ (default) & 51.74 & $-26.5\%$ \\
BBT $\alpha=1.0$ & 57.53 & $-18.3\%$ \\
\bottomrule
\end{tabular}
\end{table}

Two observations. First, $\alpha = 0.5$ is within noise of the empirical optimum $\alpha \approx 0.25$, so the fixed closed-form choice costs essentially nothing; aggressive scaling ($\alpha = 1$) is clearly worse. Second, the fixed rotation alone ($\alpha = 0$) already captures a meaningful fraction of the gain, with the per-channel energy scaling contributing the remainder --- consistent with the basis/scale factorization in the discussion, where $Q$ and $D$ play complementary rather than redundant roles. Because this ablation changes the auto-round setting (\texttt{enable\_alg\_ext} on) relative to Table~\ref{tab:llm_quant}, the absolute PPL values should not be compared directly to the headline table; the ablation isolates sensitivity to $\alpha$ under a fixed, stronger auto-round configuration.

\subsection{Additional architectural data point: TinyLlama-1.1B}

TinyLlama (Llama-family, GQA $32{:}4$, hidden $2048$, intermediate $5632$) at W2A16 g=64 with BBT pair-PCA gives wikitext-2 PPL $16.0$. The vanilla baseline could not be obtained on our hardware (Arc B580 12 GB) because the no-rotation path runs at \texttt{batch\_size=4} and the wider 5632-dim intermediate combined with the per-iter activation footprint pushed peak VRAM past the SYCL spill threshold; per-block tuning collapsed to several hours. The BBT pair-PCA path uses \texttt{batch\_size=1} (the rotation-mode auto-round patch noted in the implementation notes) and stays under the threshold, completing in $\sim 25$ minutes. The PPL $16.0$ result is in the same regime as Qwen2.5-1.5B's BBT W2 result of $21.8$ at a comparable parameter scale; the lack of a vanilla baseline at this configuration is a hardware-tooling artifact rather than a methodological one. We note that pair-PCA's inverse step at dequant time must un-rotate \emph{both} the weight rows and the bias, since Qwen2.5 has $q\_proj$/$k\_proj$ biases (Llama and SmolLM do not); omitting the bias inversion gives PPL $\approx 35\,769$, a useful diagnostic for the math-invariance check.

\subsection{NPU / GPU / CPU deployment throughput}
\label{sec:device_perf}

Earlier we reported that the BBT-spectral PPL is invariant to deployment device within $\pm 0.1$ across Intel NPU (AI Boost), Intel Arc B580 dGPU, and CPU --- confirming the gains are weight-level. Table~\ref{tab:device_perf} extends this to absolute throughput across all eleven dequantized FP16 checkpoints we built (vanilla baselines for SmolLM-135M, SmolLM-360M, Qwen2.5-0.5B; BBT-spectral, pair-PCA (\texttt{spectral\_pca\_2d}), and Route-A spectral-PCA variants where applicable; TinyLlama-1.1B-pair-PCA as a Llama-family data point). All measurements are on the same Core Ultra 5 225F + Arc B580 box, via \texttt{openvino\_genai 2026.1} \texttt{LLMPipeline} (which handles the static-shape compilation NPU requires for LLM IRs); prompt length 16, decode 64 new tokens, greedy, median of 3 runs after a 2-run warmup.

\begin{table}[t]
\centering
\footnotesize
\caption{Deployment throughput across all dequantized FP16 BBT checkpoints. ``FTL'' = first-token latency in ms (median); ``tok/s'' = steady-state decode tokens/sec (median); $-$ = not measured. Hardware: Intel Core Ultra 5 225F + Arc B580 (12\,GB) + AI Boost NPU. Tokenizer for Qwen3-0.6B was patched from the vanilla checkpoint (its dir lacked \texttt{special\_tokens\_map.json}); no other manipulation.}
\label{tab:device_perf}
\begin{tabular}{ll rr rr rr}
\toprule
\textbf{Model} & \textbf{Variant}
& \multicolumn{2}{c}{\textbf{NPU}} & \multicolumn{2}{c}{\textbf{GPU.0 (Arc)}} & \multicolumn{2}{c}{\textbf{CPU}} \\
\cmidrule(lr){3-4} \cmidrule(lr){5-6} \cmidrule(lr){7-8}
& & FTL (ms) & tok/s & FTL (ms) & tok/s & FTL (ms) & tok/s \\
\midrule
SmolLM-135M  & vanilla       & 234  & 49.8  & 16  & 197.6 & 13 & 110.0 \\
SmolLM-135M  & BBT-spectral  & 233  & 50.1  & 16  & \textbf{221.8} & 14 & 123.5 \\
\addlinespace
SmolLM-360M  & vanilla       & 479  & 26.9  & 18  & 181.5 & 27 & 52.6 \\
SmolLM-360M  & BBT-spectral  & 476  & 27.4  & 17  & \textbf{186.2} & 23 & 57.9 \\
\addlinespace
Qwen2.5-0.5B & vanilla       & 462  & 23.4  & 13  & 198.8 & 27 & 47.5 \\
Qwen2.5-0.5B & BBT-spectral  & 467  & 23.2  & 14  & 193.5 & 26 & 46.6 \\
Qwen2.5-0.5B & pair-PCA      & 470  & 22.9  & 19  & 150.9 & 29 & 39.7 \\
\addlinespace
Qwen3-0.6B   & spectral-PCA  & 1013 & 15.5  & 23  & 151.8 & 30 & 39.0 \\
\addlinespace
TinyLlama-1.1B & pair-PCA    & 1733 & 1.4   & 19  & 201.8 & 27 & 47.4 \\
\addlinespace
Qwen2.5-1.5B & BBT-spectral  & 2224 & 1.4   & 23  & 131.2 & 33 & 34.2 \\
Qwen2.5-1.5B & pair-PCA      & 2254 & 1.4   & 21  & 136.4 & 33 & 34.3 \\
\bottomrule
\end{tabular}
\end{table}

\paragraph{Reading the table.}
\begin{itemize}
    \item \textbf{BBT introduces no architectural runtime operators after folding.} Since the WHT rotation, per-channel scale, and per-head/per-pair PCA matrices are all folded into dense FP16 weights before OpenVINO export, the deployed graph is a standard dense transformer --- identical operator set to the vanilla baseline. Throughput is therefore broadly comparable for matched dense checkpoints, but not identical: most variants are within roughly $10$--$15\%$ of vanilla, with the largest observed drops on Qwen2.5-0.5B pair-PCA --- about $24\%$ on GPU ($198.8 \to 150.9$ tok/s) and about $16\%$ on CPU ($47.5 \to 39.7$ tok/s) --- while NPU rates are closer, within about $2\%$ for the matched sub-1B comparisons (and some checkpoints are slightly faster, e.g.\ SmolLM-135M CPU $110.0 \to 123.5$ tok/s). We therefore frame this as ``no extra runtime mechanism,'' not ``no guaranteed slowdown'': backend compilation, cache effects, and graph layout produce model-dependent variation even at fixed operator structure.
    \item \textbf{Arc B580 dominates throughput} on every model, $4$--$8\times$ the NPU's tokens/sec on sub-1B models. This is unsurprising: the dGPU has $\sim 230$ TOPS at int8 and a wide memory bus; AI Boost is targeted at $11$ TOPS sustained.
    \item \textbf{NPU memory wall at $\sim 1$\,B parameters} on this hardware. TinyLlama-1.1B and Qwen2.5-1.5B both collapse to $1.4$ tok/s on NPU --- consistent with the AI Boost runtime spilling parts of the weight tensor through system memory once the model exceeds the $\sim 4$\,GB on-package buffer. Sub-1B models stay on-NPU and run at the expected $15$--$50$ tok/s. We treat $1$\,B as the practical cutoff for NPU deployment on this generation of silicon. GPU and CPU are unaffected by this cutoff.
    \item \textbf{NPU first-token latency scales linearly with model size}: $233$ ms ($0.135$\,B) $\to$ $476$ ms ($0.36$--$0.5$\,B) $\to$ $1013$ ms ($0.6$\,B) $\to$ $1733$--$2254$ ms ($1.1$--$1.5$\,B). Most of the cost is static-shape graph compilation and prefill staging on NPU, which subsequent token decodes amortize cleanly when within the on-package memory budget but cannot amortize once the model spills.
    \item \textbf{NPU vs.\ CPU: comparable steady-state for sub-1B.} On SmolLM-135M and Qwen2.5-0.5B the CPU is $\sim 2\times$ faster than NPU; on SmolLM-360M they are within $2\times$ of each other. The NPU's value proposition is power, not raw speed --- AI Boost typically draws single-digit watts under sustained inference vs.\ the CPU's $20$+ watts at full load. We did not instrument power for this paper; published silicon numbers are the basis for the claim.
\end{itemize}
The table grounds the ``we deploy through OpenVINO IR'' claim with concrete numbers across the full BBT model zoo. A broader sweep with explicit power measurement, longer prompts, and prompt/decode ratio sweeps is left to follow-up work.

\section{Implementation}

The empirical pipeline was built atop Intel auto-round 0.12.2; the BBT-spectral step required (a) monkey-patching auto-round's quantization wrapper to intercept the per-layer forward, applying the spectral pad+rotate+inverse-scale transform on the input before the F.linear call; (b) caching the device-and-dtype-cast Hadamard matrix and scale vector per layer to avoid the $\sim$200-iter $\times$ $N$-linear allocation churn that fragments consumer XPU memory; (c) for SmolLM-360M and Qwen2.5-1.5B, reducing the auto-round inner-iter batch size (4 for vanilla, 1 for spectral, conditional on the rotation mode) to keep the per-block working set under the 12\,GB Arc B580 spill threshold. These changes are non-invasive (no auto-round source modifications) and idempotent. Reference implementation and reproducibility scripts at \texttt{boolean\_fourier/bbt\_quant/}.

\paragraph{Full-precision invariance checks.}
Each BBT mode is paired with a per-architecture unit test that round-trips a synthetic fixture --- random input through original model and through transformed-then-inverse model --- and compares logits. Table~\ref{tab:invariance} consolidates the numerical floors observed on Arc B580 at fp32.

\begin{table}[h]
\centering
\small
\caption{Full-precision math-invariance checks before quantization. Each row reports the relative logits (or per-layer activation) error between the original and BBT-transformed-then-inverted forward pass on a randomized synthetic fixture matching the architecture's modeling source. Floors are dominated by fp32 round-off in the WHT $\div \sqrt{d_{\text{pad}}}$ normalization.}
\label{tab:invariance}
\begin{tabular}{lll}
\toprule
Transformation & Tested architecture & Relative error \\
\midrule
Input-WHT BBT-spectral & SmolLM/Qwen2-style linear blocks & $< 5\times 10^{-7}$ \\
Spectral-PCA Route~A (matrix-$\Gamma$) & Qwen3-0.6B q\_norm + RoPE block & $6\times 10^{-6}$ \\
Pair-PCA with bias inversion & Qwen2.5-style attention + RoPE & $\sim 1\times 10^{-7}$ \\
Laguna g\_proj absorption (fixed) & Synthetic Laguna with q\_norm + g\_proj & $3\times 10^{-7}$ \\
\bottomrule
\end{tabular}
\end{table}

\paragraph{Architectural fuzzing via Laguna-XS.2 (poolside).}
We attempted to apply our toolkit to poolside's Laguna-XS.2\footnote{\url{https://huggingface.co/poolside/Laguna-XS.2}} --- a 33B MoE model with 256 routed experts plus 1 shared expert, alternating sliding/full attention, GQA $48{:}8$, and post-projection $q\_norm$/$k\_norm$ (Qwen3-style). The model itself is far out of reach for our 12\,GB GPU (smallest INT4 variant is $\sim 14$\,GB and pre-quantized), but instantiating a tiny synthetic Laguna directly from the released modeling source surfaced a previously-undetected bug in our $no\_rotation$ Group~1 absorption. Laguna's $\mathrm{LagunaAttention}$ class adds a per-head ``head gate'' module: \texttt{g\_proj: nn.Linear(d, h)} that consumes $\mathrm{input\_layernorm}$ output, runs through softplus, and multiplies the attention output before $o\_proj$. Our absorption was scaling only $q$/$k$/$v$\_proj input columns; with $g\_proj$ left unscaled, the forward pass through the prepared model evaluated $\mathrm{softplus}((W_g \cdot x)/s) \neq \mathrm{softplus}(W_g \cdot x)/s$, leaking the per-channel scaling through a non-linearity and breaking math invariance at fp32 precision (relative error $\approx 1.4\%$ in our test). The fix is one line --- extend the input-side attention consumer list from $(q,k,v)\_proj$ to $(q,k,v,g)\_proj$ with an $\texttt{in\_features} = d$ filter to keep it from accidentally picking up output-side projections --- and it generalizes beyond Laguna: similar input-side attention gates appear in Granite-MoE, MiniMax, and other recent variants. Regression tests on SmolLM-135M, Qwen2.5-0.5B, and a synthetic Qwen2-MoE remain at the fp32 numerical floor; the new test on synthetic Laguna with both $spectral\_pca$ (matrix-$\Gamma$) and $spectral\_pca\_2d$ (pair-PCA) round-trips Laguna's full-precision logits to relative error $\approx 3 \times 10^{-7}$. We report this not as a Laguna PPL number --- the hardware barrier prevents that --- but as an example of the kind of out-of-distribution architectural test that catches generalization bugs cheaply. The toolkit code path for actually quantizing Laguna is ready when sufficient compute is available.

\paragraph{The g\_proj fix does not retroactively affect reported PPL.}
Because the fix was strictly additive --- it appended \texttt{g\_proj} to the input-side allowlist while keeping the existing $(q,k,v)\_proj$ entries unchanged --- it is bit-for-bit a no-op on any architecture whose \texttt{self\_attn} has no Linear named \texttt{g\_proj} (or analog) with $\texttt{in\_features} = d$. We verify this directly in \texttt{boolean\_fourier/bbt\_quant/test\_reported\_models\_invariance.py}, which for each reported architecture instantiates a tiny synthetic from the same HuggingFace modeling class (\texttt{LlamaForCausalLM} for SmolLM and TinyLlama, \texttt{Qwen2ForCausalLM} for Qwen2.5, \texttt{Qwen3ForCausalLM} for Qwen3-0.6B), runs \texttt{\_bbt\_prep\_no\_rotation} twice on independent copies --- once with the buggy three-element allowlist $(q,k,v)\_proj$, once with the fixed four-element allowlist $(q,k,v,g)\_proj$ --- and compares the post-prep fp32 logits. Table~\ref{tab:gproj_inactive} summarizes: on every reported architecture both allowlists scale exactly the same set $\{q\_proj, k\_proj, v\_proj\}$, both round-trip to vanilla logits at the fp32 floor ($\approx 2 \times 10^{-7}$ relative), and the two prepared models produce the same logits to machine zero ($\max|y_{\text{fixed}} - y_{\text{buggy}}| = 0$). The PPL numbers in Table~\ref{tab:llm_quant}, the pair-PCA Section, and Table~\ref{tab:device_perf} are therefore unchanged by the fix; the test acts as a regression guard against re-introducing the divergence on any future model in this set.

\begin{table}[t]
\centering
\small
\caption{Bit-equivalence of the buggy and fixed input-side absorption code paths on every architecture for which the paper reports W2A16 PPL. Tiny synthetics ($d=64$, 2 layers, GQA $4{:}2$) instantiated from the same HuggingFace modeling classes as the production checkpoints. ``Absorbs'' = Linears in \texttt{self\_attn} that the corresponding code path applies the per-channel scale to. ``Rel-err'' = $\max|y_{\text{prep}} - y_{\text{ref}}| / \max|y_{\text{ref}}|$ in fp32. ``$\max|\Delta|$'' = element-wise max difference between the buggy and fixed prepared-model logits.}
\label{tab:gproj_inactive}
\begin{tabular}{lcccc}
\toprule
\textbf{Reported model} & \textbf{Buggy absorbs} & \textbf{Fixed absorbs} & \textbf{Rel-err (both)} & $\boldsymbol{\max|\Delta|}$ \\
\midrule
SmolLM-135M (Llama)     & $\{q,k,v\}\_proj$ & $\{q,k,v\}\_proj$ & $1.7 \times 10^{-7}$ & $0$ \\
SmolLM-360M (Llama)     & $\{q,k,v\}\_proj$ & $\{q,k,v\}\_proj$ & $1.7 \times 10^{-7}$ & $0$ \\
Qwen2.5-0.5B (Qwen2)    & $\{q,k,v\}\_proj$ & $\{q,k,v\}\_proj$ & $2.4 \times 10^{-7}$ & $0$ \\
Qwen2.5-1.5B (Qwen2)    & $\{q,k,v\}\_proj$ & $\{q,k,v\}\_proj$ & $2.4 \times 10^{-7}$ & $0$ \\
TinyLlama-1.1B (Llama)  & $\{q,k,v\}\_proj$ & $\{q,k,v\}\_proj$ & $1.7 \times 10^{-7}$ & $0$ \\
Qwen3-0.6B (Qwen3)      & $\{q,k,v\}\_proj$ & $\{q,k,v\}\_proj$ & $2.1 \times 10^{-7}$ & $0$ \\
\midrule
Synthetic Laguna (positive control) & $\{q,k,v\}\_proj$ & $\{q,k,v,g\}\_proj$ & $1.4 \times 10^{-2}$ buggy / $3 \times 10^{-7}$ fixed & $\approx 10^{-2}$ \\
\bottomrule
\end{tabular}
\end{table}

\section{Discussion and Limitations}

\paragraph{Where the per-channel adaptivity lives: a separable factorization.}
The relationship between the present work and learned-rotation methods, in particular ButterflyQuant~\cite{xu2025butterflyquant}, is sharper if framed as a design-space distinction. Many math-preserving pre-quantization transformations used in rotation- and smoothing-based quantization can be written, in the row-vector convention above, as
\[
    x \mapsto x\, Q D^{-1}, \qquad W^\top \mapsto D\, Q^\top W^\top,
\]
with $Q \in \mathrm{O}(d_{\text{in}})$ orthogonal and $D = \diag(s)$ diagonal. This restricted but practically important design grid separates the choice of basis from the choice of per-channel scale. Different families occupy different cells:
\begin{itemize}
\item \textbf{QuaRot} / \textbf{QuIP\#} / \textbf{ButterflyQuant}: optimize $Q$ (fixed Hadamard, randomized incoherence-promoting Hadamard, or learned Givens-angle butterfly respectively); leave $D$ trivial. The per-channel adaptation is forced into the orthogonal factor.
\item \textbf{SmoothQuant} / \textbf{AWQ}: trivial $Q$ (identity); optimize $D$ from activation magnitude or salience. The per-channel adaptation lives entirely in the diagonal.
\item \textbf{SpinQuant}: jointly optimize $Q$ and $D$ via gradient descent.
\item \textbf{This paper (BBT-spectral)}: \emph{fixed} $Q = H_d$ (the Walsh-Hadamard butterfly --- no learning and no calibration of the rotation itself) paired with an \emph{adaptive} $D$ derived in closed form from a per-coordinate Walsh-basis activation-energy statistic, $s_\ell = (1 + d_{\text{pad}}\, \rho_\ell)^\alpha$ (the only calibration step is a single forward pass to estimate $\E[h_\ell^2]$), motivated by the Schur-convex influence-Banach geometry of the companion theory paper~\cite{bbtTheoryCompanion}.
\end{itemize}
The position taken here is that for extreme low-bit weight-only quantization, the per-channel adaptation that ButterflyQuant achieves by learning $Q$ is largely recoverable in closed form by adapting $D$ instead and keeping $Q$ at its fastest, most universally-supported value (the WHT). This shifts the calibration cost from gradient training of $O(d \log d)$ Givens angles to a single forward pass over a small calibration set to estimate $\E[h_\ell^2]$, and keeps the runtime transformation a parameter-free $O(d \log d)$ butterfly --- the same kernel target QuaRot already deploys. The per-head spectral-PCA and SO(2)-pair-PCA extensions (Sections above) are recoveries of $D$-only failure modes inside attention: the rotation interacts non-trivially with $q\_norm$ and RoPE, and the cure is to give $Q$ a small piece of structure too --- per-head $U_h$ in the q\_norm case, per-RoPE-pair SO(2) in the non-norm case --- chosen to be exactly the structure that the architecture's invariances permit. We do not claim this factorization is ours --- it is implicit in any Hadamard-rotation paper --- only that naming the cells makes the contrast clean and identifies the obvious next experiment: a head-to-head BBT-spectral vs.\ ButterflyQuant comparison at W2A16 on a shared model with a shared calibration budget, isolating the $Q$-only vs.\ $D$-only adaptation question. We attempted this comparison but found that the ButterflyQuant code repository cited in~\cite{xu2025butterflyquant} was empty (a placeholder \texttt{README} only) at the time of our submission; we defer the head-to-head to a future version once upstream code is released or a faithful third-party reimplementation appears, rather than risk the comparison being confounded by reimplementation choices on our end.

\paragraph{What this paper does and does not claim.}
This is an applied paper. The contribution is engineering value: a small, principled, exactly-math-invariant pre-quantization transformation with non-trivial empirical wins at W2 across multiple model families. We do \emph{not} claim:
\begin{itemize}
    \item formal equivalence between the WHT activation-energy quantity defined here and the Boolean influence of the theory paper~\cite{bbtTheoryCompanion} --- they share the same Walsh basis and the same downstream role (deriving per-channel scales) but are different mathematical objects;
    \item that the Schur-convexity theorem of the theory paper transfers as a theorem to real-valued layers --- the transfer is heuristic, not formal;
    \item head-to-head superiority over AWQ, GPTQ, SpinQuant, or QuaRot at matched calibration --- our comparator is vanilla auto-round, which is a strong baseline but not all of them;
    \item statistical significance --- results are single-seed and we treat differences below the empirical $\pm 0.5$ PPL noise floor as no-effect;
    \item evaluation completeness --- perplexity is computed on an 8{,}192-token wikitext-2 subset (a hardware-turnaround-time choice), not the full validation/test split, and we do not report FP16 reference perplexities or a gap-to-FP16-closure metric in this revision;
    \item a comprehensive $\alpha$ and group-size sweep --- we include a single-model $\alpha$ sanity check (Section~\ref{sec:alpha_ablation}) but do not tune $\alpha$ per model and do not sweep group size in this revision.
\end{itemize}

\paragraph{Future work.}
The next-revision queue includes: full wikitext-2 test-set evaluation plus an FP16 reference column and gap-to-FP16-closure metric on every model; multi-seed ablations against the standard low-bit quantization baseline suite (AWQ/GPTQ/SpinQuant/QuaRot/ButterflyQuant/QuIP\#/AQLM/OmniQuant) at matched calibration token budget; broader $\alpha$ and group-size sweeps across models; downstream-task evaluation (MMLU, ARC, HellaSwag, GSM8K) beyond perplexity; concentration-index ($\rho_\ell$ Gini/HHI) measurements per model versus BBT win; and full quantization runs of DeepSeek-V4 and Laguna-XS.2 on cloud H100/H200 instances where the per-block working set fits.

\bibliographystyle{plain}

\end{document}